# PDCNet: a benchmark and general deep learning framework for activity prediction of peptide-drug conjugates

Yun Liu[1,#], Jintu Huang[1,#], Yingying Zhu[1,#] , Congrui Wen[1], Yu Pang[1], Ji-Quan Zhang[2,*] and Ling Wang[1,*]

[1]*Joint International Research Laboratory of Synthetic Biology and Medicine, Ministry of Education, Guangdong Provincial Key Laboratory of Fermentation and Enzyme Engineering, Guangdong Provincial Engineering and Technology Research Center of Biopharmaceuticals, School of Biology and Biological Engineering, South China University of Technology, Guangzhou 510006, China.*

*State Key Laboratory of Discovery and Utilization of Functional Components in* [2]*Traditional Chinese Medicine, Guizhou Provincial Engineering Technology Research Center for Chemical Drug R&D, School of Pharmaceutical Sciences, Guizhou Medical University, Guiyang 561113, P.R. China.*

[#]*These authors contributed equally.*

**Corresponding author: lingwang@scut.edu.cn (L. Wang), zjqgmc@163.com (J. Zhang)*

## Abstract

Peptide-drug conjugates (PDCs) represent a promising therapeutic avenue for human diseases, particularly in cancer treatment. Systematic elucidation of structure-activity relationships (SARs) and accurate prediction of the activity of PDCs are critical for the rational design and optimization of these conjugates. To this end, we carefully design and construct a benchmark PDCs dataset compiled from literature-derived collections and PDCdb database, and then develop PDCNet, the first unified deep learning framework for forecasting the activity of PDCs. The architecture systematically captures the complex factors underlying anticancer decisions of PDCs in real-word scenarios through a multi-level feature fusion framework that collaboratively characterizes and learns the features of peptides, linkers, and payloads. Leveraging a curated PDCs benchmark dataset, comprehensive evaluation results show that PDCNet demonstrates superior predictive capability, with the highest AUC, F1, MCC and BA scores of 0.9213, 0.7656, 0.7071 and 0.8388 for the test set, outperforming eight established traditional machine learning models. Multi-level validations, including 5-fold cross-validation, threshold testing, ablation studies, model interpretability analysis and external independent testing, further confirm the superiority, robustness, and usability of the PDCNet architecture. We anticipate that PDCNet represents a novel paradigm, incorporating both a benchmark dataset and advanced models, which can accelerate the design and discovery of new PDC-based therapeutic agents.

## Introduction

The concept of “magic bullets”, first proposed by Paul Ehrlich in 1913, represents a ground-breaking therapeutic paradigm aimed at delivering cytotoxic agents to cancer cells with high specificity. This innovative approach envisions the use of targeted carriers to guide potent drugs to malignant cells, thereby minimizing damage to healthy tissues[1, 2]. In accordance with this pioneering vision, antibody-drug conjugates (ADCs) have emerged as the first practical realization of this concept[3, 4]. ADCs are intricate molecular constructs comprising three essential components: a monoclonal antibody that selectively binds to specific cancer cell surface antigens, a stable chemical linker that connects the antibody to the payload, and a highly potent cytotoxic agent[5]. Despite the remarkable achievements of ADCs in the field of cancer treatment, they still face limitations stemming from the large molecular size of antibodies, poor tumor penetration, high immunogenicity, and high production costs[6, 7, 8].

Peptide-drug conjugates (PDCs) have emerged as a transformative class of targeted oncology therapeutics, increasingly recognized as next-generation successors to ADCs [9]. Structurally analogous to ADCs, PDCs are macromolecular systems that conjugate cytotoxic agents to tumor-targeting peptides or cell-penetrating peptides through the use of appropriate linkers[10]. While maintaining architectural similarities with ADCs in their linker-mediated drug-carrier configuration, PDCs exhibit a distinct molecular weight profile of 1-5 kDa, contrasting sharply with the 160 kDa antibody framework characteristic of ADCs. Peptides, as integral components of proteins, exhibit selectivity, reduced immunogenicity, and targeted action[11]. Moreover, the high receptor specificity of targeting peptides also aids in mitigating the off-target toxicity associated with small-molecule drugs[12]. The mechanism of PDCs is illustrated in Fig. 1a. PDCs usually conjugate to cell-penetrating peptides to enter cancer cells via transmembrane-mediated internalization, whereas those conjugated to cell-targeting peptides enter via receptor-mediated endocytosis[13]. PDCs demonstrate enhanced anticancer efficacy comparable to conventional small-molecule drugs. For example, TH1902, as a novel docetaxel-peptide conjugate targeting the Sortilin (SORT1) receptor[14], showed enhanced apoptotic effects compared to free docetaxel. Currently, 17 candidates currently in clinical development demonstrate advantages in cancer treatment. However, the stability and targeting efficiency of peptides, the ability of linkers to enable payload release, and the potent cytotoxicity of the payloads all pose considerable challenges in the development of PDCs[10, 15].

Compared to small-molecule drugs, the rational computational design of peptide-drug conjugates (PDCs) remains underexplored due to their structural complexity, with current development predominantly relying on empirical approaches. Recent advancements demonstrate the potential of computational strategies in targeted PDC design. For instance, Zhang et al. utilized molecular dynamics simulations and binding free energy calculations to design ROR1-targeting peptide mimetics, followed by computer-aided virtual mutation analysis to optimize the peptide sequences[16]. This approach yielded peptides with nanomolar-level affinity, and enabled the construction of PDCs, among which compound II-3 exhibited remarkable antitumor activity against MDA-MB-231 cells and pharmacokinetic properties, validating computational efficacy in PDC design. Similar, Muratspahić et al. integrated Rosetta-based protein design, cryo-EM structures, and molecular dynamics to develop a κ-opioid receptor (KOR)-targeted PDC[17]. Their optimized conjugate, DNCP-β-NalA (1), exhibited high KOR affinity, selectivity, and G-protein-biased activation, mitigating β-arrestin-related side effects. *In vivo* studies confirmed its analgesic efficacy without motor impairment or sedation. Despite these successes, the field lacks robust computational tools to systematically elucidate structure-activity relationships (SARs) and predict PDCs activity. Current methods, while effective for specific targets, remain fragmented and lack generalizability, hindering the development of unified frameworks for rational PDC design and discovery. Although deep learning has made significant progress in predicting molecular properties[18, 19, 20, 21, 22], discovering active compounds[23, 24], and generating molecules[25], to the best of our knowledge, there have been no reports on the application of deep learning technologies to the development of PDC drugs.

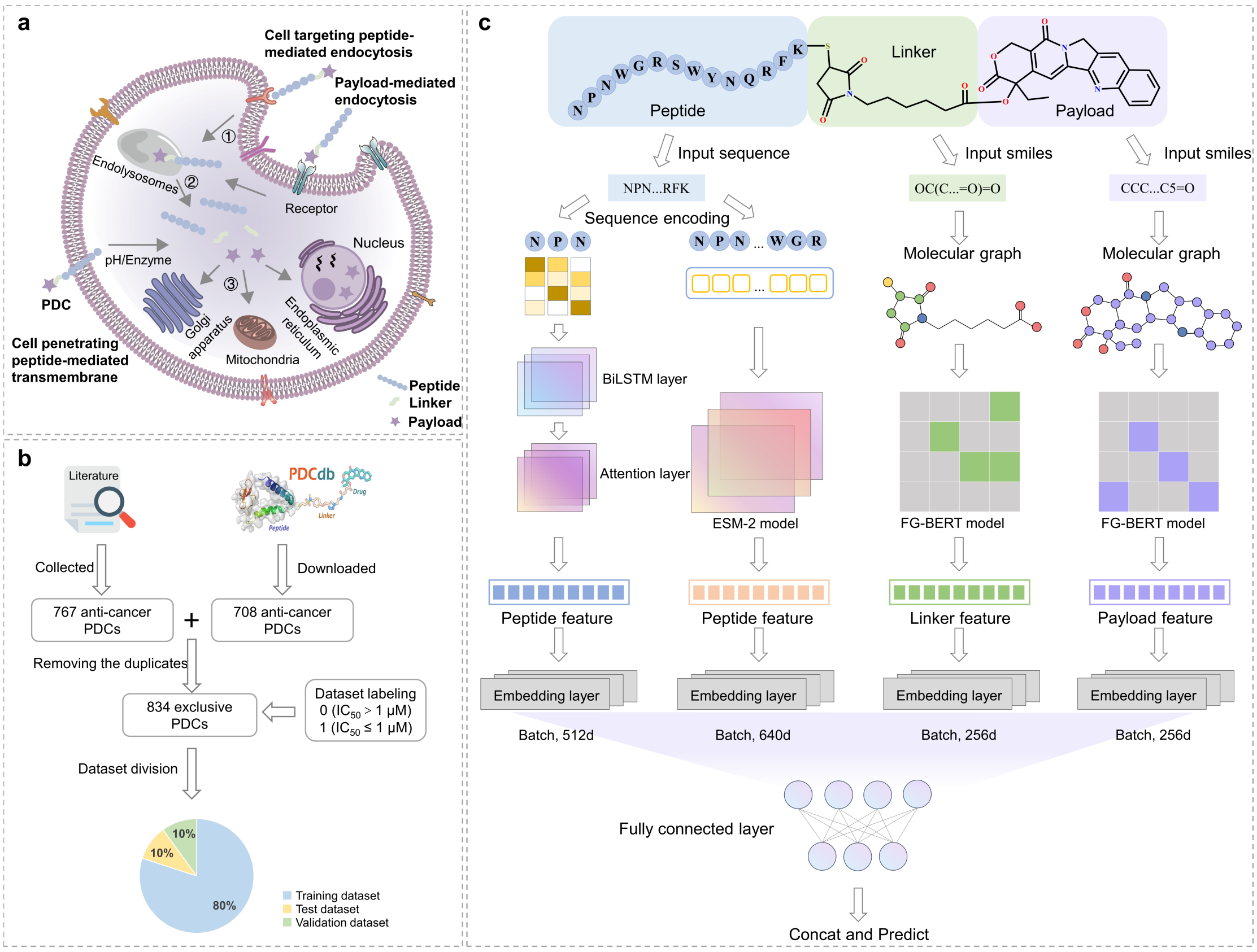

**Fig. 1 | A schematic diagram of the PDC benchmark dataset and architecture involved in the PDCNet model. a**, Mechanism of action of PDCs. **b**, Construction of the PDC benchmark dataset. **c**, The network architecture of PDCNet model.

To address these challenges, we first constructed a benchmark dataset (Fig. 1b) comprising PDC structures and their activity labels, which exhibits strong structural diversity and broad coverage of chemical space. Based on this benchmark dataset, we developed PDCNet (Fig. 1c), a deep learning-based multimodal framework for PDC activity prediction. Subsequently, we conducted a comprehensive evaluation of the model's architecture advanced nature across multiple levels, including 5-fold cross-validation, ablation studies, threshold reclassification experiments, t-SNE dimensionality reduction visualization, independent external dataset validation, and interpretability analysis. These evaluation results demonstrate the superior performance of the PDCNet model and its practical utility in real-word scenarios.

## Results

### PDCs benchmark dataset analysis and visualization

To date, no standardized dataset has existed in the PDC field for exploring structure-activity relationships and activity prediction. To address this gap, we established a standardized workflow (Fig. 1b) and constructed the first benchmark PDC dataset. Following rigorous standardization, we developed a modeling-ready dataset comprising 834 anticancer PDCs. Each entry incorporates three key components: the peptide sequence, linker SMILES, and payload SMILES (Fig. 1c). Among these, 209 PDCs were labelled as "1" (i.e. "active"), whereas 625 PDCs were labelled as "0" (i.e. "inactive"). As shown in Fig. 2a, the number of unique peptides, linkers, and payloads are 408, 231, and 202, respectively, indicating the structural diversity of PDCs in the dataset. Dataset analysis revealed 2 clinically approved PDCs, 16 in clinical trials, 1 undergoing preclinical development, and the remainder (815) in exploratory research phases (Fig. 2b). Given that the characteristics of three components (peptide, linker and payload) in a PDC play a critical role in maintaining their activities, we analyzed the distributions of these PDC components in this dataset.

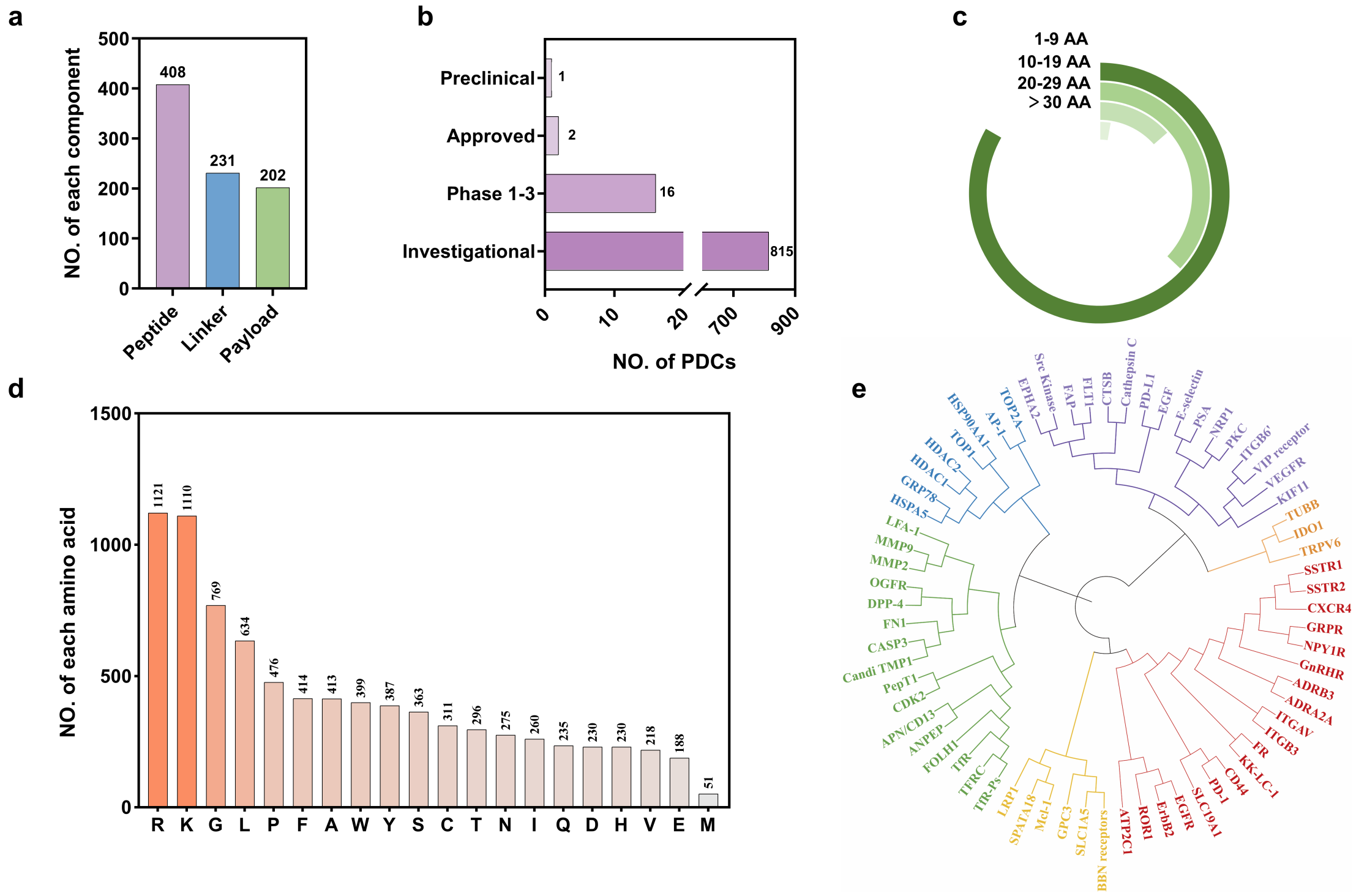

**Fig. 2 │ Visualization and analysis of peptide items in the PDC benchmark dataset.** **a**, The quantity statistics of peptides, linkers and payloads in the PDC dataset. **b**, Distribution of research status for PDCs in the dataset. **c**, Distribution of the lengths of peptides in the dataset. **d**, The quantity statistics of each amino acid in the data set. **e**, Developmental tree analysis of targets in the dataset. The developmental tree was generated by MEGA11 software (https://www.megasoftware.net/)[26, 27].

As peptide length is a critical factor influencing the stability of PDCs and cell membrane permeability[28, 29], the distribution of peptide lengths (Fig. 2c) illustrates that the majority of PDCs have peptides of 20 amino acids or fewer. Specifically, the number of PDCs within 10 amino acids reaches 508. Fig. 2d illustrates the frequency of each of the 20 standard amino acids, demonstrating that Arg, Lys, and Gly are the most frequently occurring amino acids in these peptide items. We further analyzed the receptors targeted by the PDCs in this dataset as shown in Fig. 2e. These receptors can be divided into three main categories, of which the top three most common are Gonadotropin-releasing hormone receptor (GnRHR), receptor tyrosine-protein kinase erbB-2 (ErbB2) and somatostatin receptor type 2 (SSTR2).

In PDC design, the payload serves as the therapeutic core whose rational selection dictates targeting precision and safety. As shown in Fig. 3a and Fig. 3b, ten payloads including doxorubicin, daunorubicin, campathecin are widely used to design various PDCs for potential cancer treatment, highlighting that the predominant payload types in the dataset are highly cytotoxic drugs, consistent with the established design paradigm for PDCs[30].

In PDCs, linkers chemically connect the peptide carrier to the therapeutic payload. They ensure stability during systemic circulation, prevent premature drug release, and enable controlled activation at the target site (e.g., via enzymatic cleavage, pH sensitivity, or redox conditions), optimizing drug delivery, reducing off-target toxicity, and enhancing therapeutic efficacy. To characterize these linkers, we classified them into twelve structural categories based on similarity. Fig. 3c presents the distribution of linkers across these categories, while Fig. 3d shows the most frequently occurring types, including amide bond, succinimidyl thioether bond, ester bond, disulfide bond, and glutaryl linkers. These results highlight the extensive structural diversity of linkers in the PDC modeling dataset.

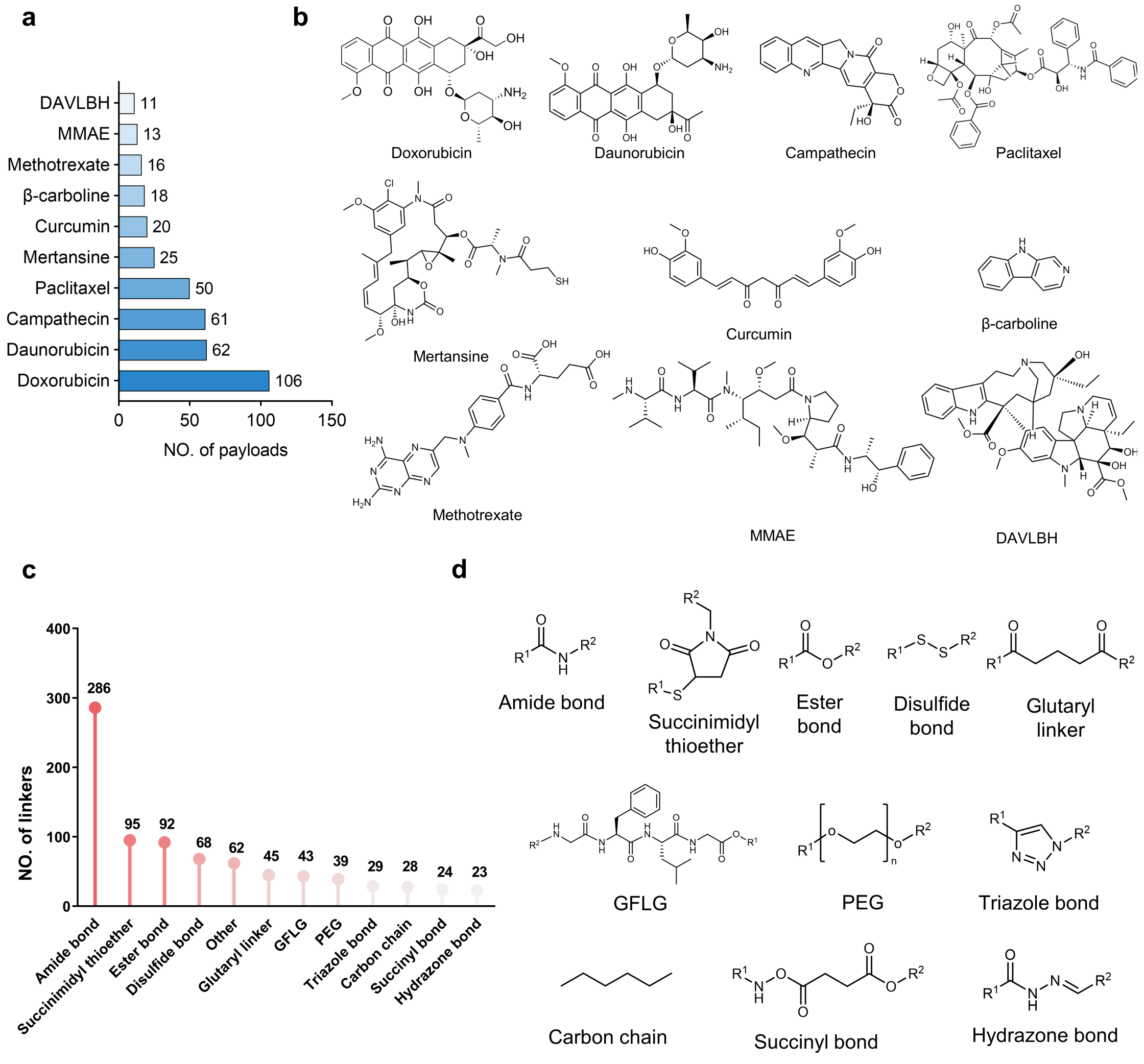

e

| | No. of compounds | No. of scaffolds | Scaffolds/compounds (%) |
|---|---|---|---|
| **Payloads** | **202** | **114** | **56.44** |
| **Linkers** | **231** | **27** | **11.69** |

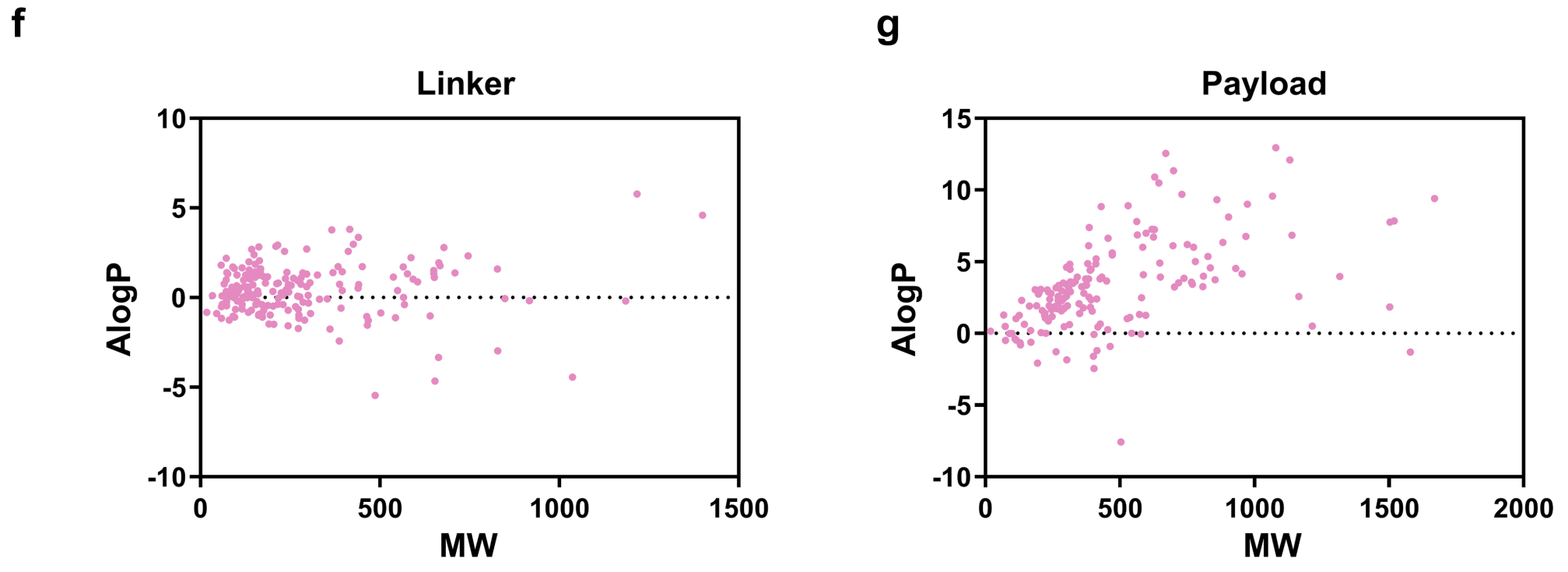

**Fig. 3 | Visualization and analysis of linkers and payloads in the PDC benchmark dataset. a**, Distribution of top 10 payloads in PDC design. **b**, The structures of the top

payloads. **c**, Clustering and distribution analysis of linkers. **d**, The structures of the top linkers. **e**, Scaffolds diversity of payloads and linkers in the dataset. **f**, The chemical space analysis of linkers and payloads in the PDC benchmark dataset. The AlogP and MW were computed by RDKit[31].

Furthermore, we found that in the PDCNet modeling dataset, the proportions of linkers and payload scaffolds were 11.69% and 56.44%, respectively (Fig. 3e). This distribution may reflect the current status of PDC design: during PDC structural optimization, researchers tend to modify the structures of linkers to improve release efficiency, while making relatively fewer changes to payload scaffolds. This is because the primary role of linkers is to stabilize conjugation and control release, and modifying linkers is more likely to generate novel PDC structures. Modifications to payloads often cause them to deviate from the structures of marketed drugs, which may lead to uncertain additional toxicity issues. In addition, the AlogP and molecular weight (MW) of the linkers and payloads were computed to analyze the chemical space of the PDCs modelling dataset. Fig. 3f and 3g reveal that the chemical space spanned by the linkers (MW: 18.015–1399.509; AlogP: -5.458–5.767) and payloads (MW: 19.009–1669.615; AlogP: -7.573–12.949) in the dataset is expansive, further demonstrating the chemical diversity of the dataset. Collectively, the PDC modelling dataset constructed in this study exhibits a wide range of chemical space, extensive structural diversity, and compositional complexity, which poses challenges to computational methods in predicting the anticancer activities of PDCs. Of course, computational models built based on this dataset can not only fully simulate PDCs in the real world, but also enhance the robustness and accuracy of the model in predicting the activity of this class of drugs.

**Architecture and performance of PDCNet**

Although PDCs represent a critically important class of therapeutic agents, particularly in the field of cancer treatment, their current design and discovery remain predominantly experience-driven. To the best of our knowledge, no computational methods or predictive models have been reported to data that elucidate the complex SARs of PDC and predict their activity. In this study, we developed the PDCNet framework (Fig. 1c) that systematically learns the distinct characteristics of three PDC components (peptide, linker, and payload) to attempt accurate explore the SARs of PDCs and activity prediction. Furthermore, we established traditional machine learning approaches as baseline models for predictive performance comparison with PDCNet. Notably, despite being considered baseline models, they represent the first computational models for PDC activity prediction. For fair comparison, all models were trained and tested with three different random seeds, maintaining the same data split proportion and method. The average of the three runs was adopted as the final evaluation metrics for model prediction performance.

The performance metrics of PDCNet and eight baseline models are presented in Table 1 for the test set. As shown Table 1, our PDCNet model demonstrates superior predictive performance compared to eight baselines. For one thing, given the pronounced class imbalance on the PDCs modelling dataset, we pay particular attention to the four key metrics of F1, AUC, MCC and BA. First, it is evident that PDCNet achieves the highest F1 score (0.7656), demonstrating its effectiveness in predicting minority class samples. Additionally, PDCNet also achieves the highest AUC (0.9213), dictating its capacity to effectively discriminate between positive and negative samples, while concurrently managing false positives and false negatives in a balanced manner. This ability is of particular significance when dealing with imbalanced datasets. Further analytical results demonstrate that PDCNet achieves 38.46% and 23.83% improvements in AUC compared to the weakest and strongest baseline models (RF_Morgan and LR_Morgan), respectively. This substantial performance gap unequivocally confirms the architectural superiority of PDCNet in addressing the compositional complexity inherent to PDCs. The F1 score and AUC are complementary metrics, and the excellent performance of PDCNet in both of them highlights its outstanding ability to distinguish between different classes in the imbalanced PDCs dataset and its capacity to accurately identify the minority classes. Meanwhile, PDCNet exhibits robust stability, as evidenced by achieving the highest scores on MCC (0.7071) and BA (0.8388), which are well-suited for comprehensively evaluating the performance of this model on imbalanced dataset[32, 33].

For another, PDCNet attained the highest values on PRAUC (0.7363), PPV (0.8131), and NPV (0.9250), which collectively indicate that PDCNet can effectively minimize both false positives and false negatives, thereby ensuring the reliability of its predictive outcomes. Furthermore, the high SE (0.7239) of the PDCNet further underscores its ability to accurately identify true positives, representing the most substantial improvement over baseline models, which achieved SE values in the range of 0.38–0.56. Notably, PDCNet and all baseline models achieved remarkably consistent high specificity (SP) values (range: 0.9084–0.9593), demonstrating robust performance in identifying true negative samples. This observation aligns with prior studies indicating that class-imbalanced datasets inherently favor high specificity when negative samples dominate[34]. Crucially, PDCNet maintained this baseline-level SP while simultaneously achieving the aforementioned sensitivity (SE) improvements, demonstrating its a dual capability critical for therapeutic applications requiring both precision in minority-class detection (e.g., active PDCs) and reliability in majority-class exclusion (e.g., inactive PDCs). Furthermore, PDCNet demonstrates superior overall predictive performance with the highest ACC value of 0.9036. Collectively, these findings not only demonstrate the superior predictive performance of the PDCNet model but also indirectly illustrate the advanced nature of PDCNet architecture, which enables it to better handle the inherent complexity of PDC drugs.

**Table 1 │ Performance results of PDCNet and traditional ML-based models on the test set.**

| Model | SE | SP | MCC | ACC | AUC | F1 | BA | PRAUC | PPV | NPV |
|---|---|---|---|---|---|---|---|---|---|---|
| PDCNet | **0.7239 ± 0.0263** | 0.9538 ± 0.0104 | **0.7071 ± 0.0383** | **0.9036 ± 0.0085** | **0.9213 ± 0.0273** | **0.7656 ± 0.0338** | **0.8388 ± 0.0162** | **0.7363 ± 0.1181** | **0.8131 ± 0.0470** | **0.9250 ± 0.0016** |
| RF_Morgan | 0.3817 ± 0.0444 | 0.9490 ± 0.0129 | 0.4159 ± 0.0374 | 0.8175 ± 0.0148 | 0.6654 ± 0.0193 | 0.4875 ± 0.0363 | 0.6654 ± 0.0193 | 0.6067 ± 0.0499 | 0.6889 ± 0.0831 | 0.8366 ± 0.0281 |
| RF_MACCS | 0.4252 ± 0.0529 | 0.9544 ± 0.0206 | 0.4675 ± 0.0926 | 0.8333 ± 0.0097 | 0.6898 ± 0.0345 | 0.5355 ± 0.0752 | 0.6898 ± 0.0345 | 0.6420 ± 0.0946 | 0.7278 ± 0.1363 | 0.8486 ± 0.0115 |
| SVM_Morgan | 0.5171 ± 0.0125 | 0.9336 ± 0.0170 | 0.5034 ± 0.0364 | 0.8373 ± 0.0056 | 0.7253 ± 0.0078 | 0.5916 ± 0.0292 | 0.7253 ± 0.0078 | 0.6628 ± 0.0475 | 0.6974 ± 0.0817 | 0.8660 ± 0.0185 |
| SVM_MACCS | 0.5356 ± 0.0145 | 0.9390 ± 0.0232 | 0.5305 ± 0.0515 | 0.8452 ± 0.0097 | 0.7373 ± 0.0123 | 0.6121 ± 0.0363 | 0.7373 ± 0.0123 | 0.6833 ± 0.0582 | 0.7238 ± 0.1052 | 0.8709 ± 0.0198 |
| LR_Morgan | 0.5646 ± 0.0330 | 0.9235 ± 0.0194 | 0.5227 ± 0.0702 | 0.8413 ± 0.0112 | 0.7440 ± 0.0258 | 0.6169 ± 0.0597 | 0.7440 ± 0.0258 | 0.6733 ± 0.0689 | 0.6828 ± 0.1000 | 0.8774 ± 0.0083 |
| LR_MACCS | 0.5449 ± 0.0569 | 0.9084 ± 0.0308 | 0.4798 ± 0.1019 | 0.8254 ± 0.0224 | 0.7267 ± 0.0400 | 0.5852 ± 0.0843 | 0.7267 ± 0.0400 | 0.6428 ± 0.0918 | 0.6375 ± 0.1319 | 0.8713 ± 0.0055 |
| XGB_Morgan | 0.5200 ± 0.0645 | 0.9449 ± 0.0394 | 0.5341 ± 0.0921 | 0.8452 ± 0.0168 | 0.7324 ± 0.0319 | 0.6034 ± 0.0617 | 0.7324 ± 0.0319 | 0.6920 ± 0.0963 | 0.7530 ± 0.1810 | 0.8680 ± 0.0252 |
| XGB_MACCS | 0.4767 ± 0.0886 | **0.9593 ± 0.0355** | 0.5333 ± 0.0635 | 0.8492 ± 0.0112 | 0.7180 ± 0.0356 | 0.5850 ± 0.0681 | 0.7180 ± 0.0356 | 0.6988 ± 0.0644 | 0.8017 ± 0.1278 | 0.8615 ± 0.0155 |

All models are trained and tested with three random seeds (1~3), the same data split proportion and split method are utilized. The best results are highlighted in bold.

### 5-fold cross-validation for PDCNet model

To evaluate the reliability and stability of the PDCNet model, we employed 5-fold cross-validation, a method particularly suitable for addressing classification imbalance and small to medium sized datasets. As shown in Fig. 4a, the outcomes of cross-validation aligned consistently with those from standard training across all evaluation metrics, indicating that PDCNet possesses excellent generalization capabilities and stability. With regard to the characteristics of the PDCNet benchmark dataset, we focus on the four metrics: BA, MCC, AUC and ACC. Specifically, the average MCC, BA, ACC and AUC of the 5-fold cross-validation are 0.7840 ± 0.0779, 0.8794 ± 0.0536, 0.9194 ± 0.0308 and 0.9464 ± 0.0572, respectively. These results showed negligible differences compared to these obtained from the previously trained PDCNet model. In summary, the 5-fold cross-validation demonstrated that the PDCNet model has robust stability and generalization ability.

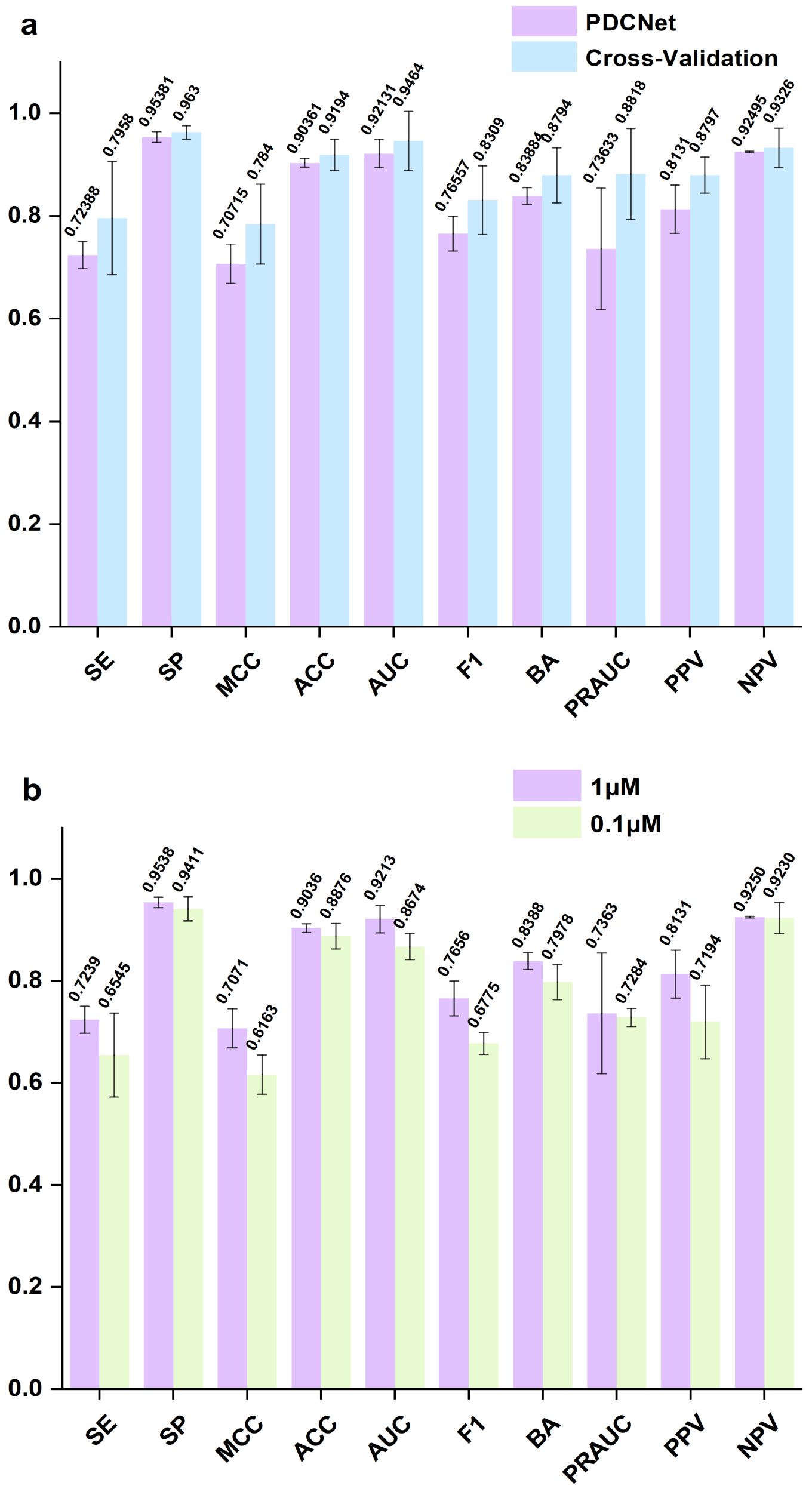

**Fig. 4 | Validation results of PDCNet model. a**, Comparison results of the original PDCNet model with 5-fold cross-validation model. **b**, Comparison results of the original PDCNet model with the model based on a threshold of 0.1 μM. The average metrics of three independent experiments conducted on the test set were used to evaluate all models.

**Activity threshold-based evaluation confirms the efficacy of PDCNet**

With the aim of examining the robustness and generalization ability of our model, we further adopted another threshold to judge the activity/inactivity of the PDCs in the benchmark dataset, Concretely, we now labelled PDCs with $IC_{50}$, $EC_{50}$ and $GI_{50}$ below 0.1 μM in the cell viability assays as "1" (active), and otherwise as "0" (inactive). PDCs that have been tested in animals, entered clinical trials, or have been marketed are still labelled as active. After retraining, Fig. 4b shows that there is not much difference between the model training results under two different activity judgement thresholds. For example, the values of MCC, AUC, PRAUC, and F1 after retraining are 0.6163, 0.8674, 0.6775, and 0.7284, respectively, which are consistent with the performance of the results (0.7071, 0.9213, 0.7363, 0.7656) obtained by training with the initial activity judgement thresholds. This observation signifies that PDCNet not only exhibits notable robustness and generalization capability, but also has a strong anti-interference ability to potential noise or boundary samples in the data. Considering the characteristics of the existing PDCs benchmark dataset, we propose that selecting an activity threshold of 1 μM is scientifically justified.

**Ablation experiments**

As shown in Fig. 1c, PDCNet is a multipath architecture that can learn and combine useful information of the peptides, linkers, and payloads involved in the PDCs. To further investigate the necessity of these three well-designed feature extraction modules, we conducted ablation experiments by designing five variants of PDCNet: 1) PDCNet without peptide embedding information (w/o embed) in peptide feature extraction module; 2) PDCNet without peptide encoding information (w/o encode) in peptide feature extraction module; 3) PDCNet without peptide information (w/o peptide) in peptide feature extraction module; 4) PDCNet without linker information (w/o linker) in linker feature extraction module ; and 5) PDCNet without payload information (w/o payload) in payload feature extraction module. The five variants of PDCNet were rigorously evaluated on the standardized PDC benchmark dataset, maintaining identical experimental configurations with the original PDCNet framework to ensure equitable comparative analysis.

As shown in Fig. 5, all five architectural variants of PDCNet exhibited varying degrees of performance degradation across four critical metrics (F1, BA, AUC, and MCC). Notably, the exclusion of payload information resulted in the most severe metric deterioration, rendering the model nearly inoperable. Specifically, F1 score plummeted from 0.7656 to 0.496 (35.21% reduction, Fig. 5a), BA score declined from 0.8388 to 0.6741 (19.64% decrease, Fig. 5b), AUC decreased from 0.9213 to 0.8169 (11.33% decline, Fig. 5c), while MCC dropped from 0.7071 to 0.44 (37.77% reduction, Fig. 5d). This phenomenon aligns with the fundamental pharmacological principle that payloads in PDCs serve as the critical determinant for tumor cell inhibition, where optimal payload selection is indispensable for designing efficacious PDCs. Secondarily, linker presence significantly impacted model performance (Fig. 5), as linkers fulfill dual mechanistic roles of bridging peptides and payloads structurally, and enabling tumor microenvironment-responsive payload release through cleavable bonds, thereby establishing linkers as non-negotiable components in PDCs. The absence of peptide information, that is, the model's performance when making predictions without any input of peptide sequences, shows a smaller decline compared to the former two architectural variants. This may be because the peptide sequence in PDC design primarily serves the targeting recognition function, and its structure itself is not the core determinant of molecular activity. Furthermore, the dual-channel peptide encoding in PDCNet demonstrated superior feature extraction capabilities compared to single-channel peptide encoding, directly enhancing predictive performance. Similar degradation trends were observed in other metrics (Fig. S1), including ACC and SE. Collectively, these ablation study results scientifically validate the necessity and superiority of PDCNet architecture.

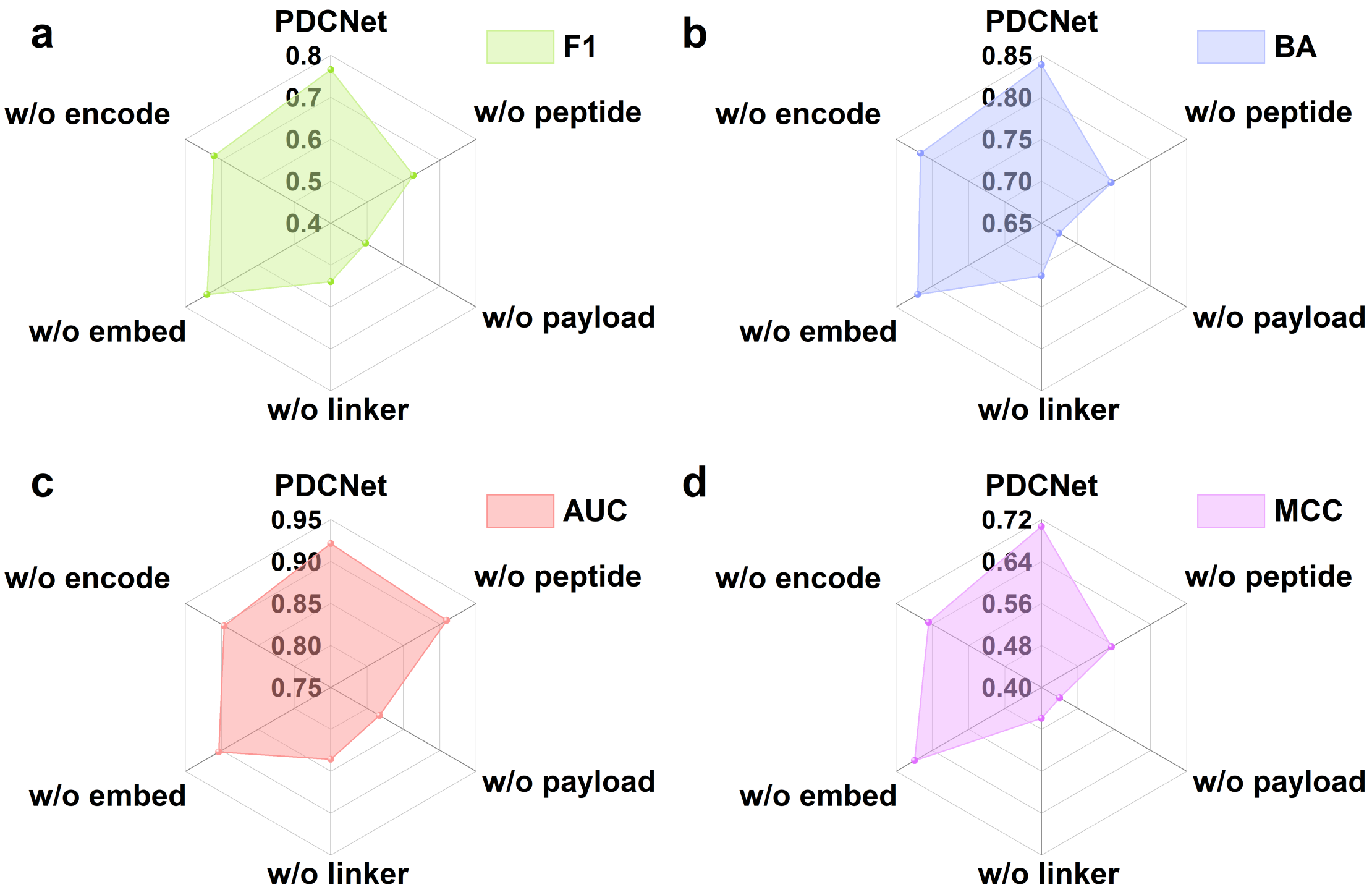

**Fig. 5 | Results of ablation experiments. a**, **b**, **c** and **d** represent the average F1, BA, AUC and MCC values of the test set, respectively.

**t-SNE dimensionality reduction visualization of PDCNet**

To evaluate whether the PDCNet effectively learns discriminative PDC feature representations, we employed t-SNE[35] to visualize the high-dimensional features extracted by the model before and after training. Specifically, we reduced the features extracted by the PDCNet model on the PDC dataset to a two-dimensional space and compared the embedding distributions before and after training to observe the evolution of feature representation. As shown in Fig. 6a, before training, the active and inactive PDC samples were distributed in a messy and overlapping manner in the two-dimensional space, with no clear class boundaries. This indicates that the untrained model was unable to distinguish the fused features. After training, the t-SNE results revealed a clearer trend of class clustering, with active and inactive samples forming separate distribution regions. This demonstrates that the model has successfully learned features related to activity and has developed strong discriminative capabilities.

To further analyze the model's learning ability for samples with different structures, we selected two pairs of representative PDC samples for comparison. As shown in Fig. 6b, the first pair of samples (PDC1 and PDC2) had highly similar structures. Before training, they were almost overlapping in the feature space, but after training, they were successfully distinguished and classified into different activity regions. This indicates that the model can capture subtle but critical structural differences and map them into deep features. Fig. 6c shows the second pair of samples (PDC3 and PDC4), which had significant structural differences. Before training, they were distributed in a discrete and irregular manner, but after training, they were accurately classified, further verifying the robustness and generalization ability of PDCNet in handling structurally heterogeneous samples.

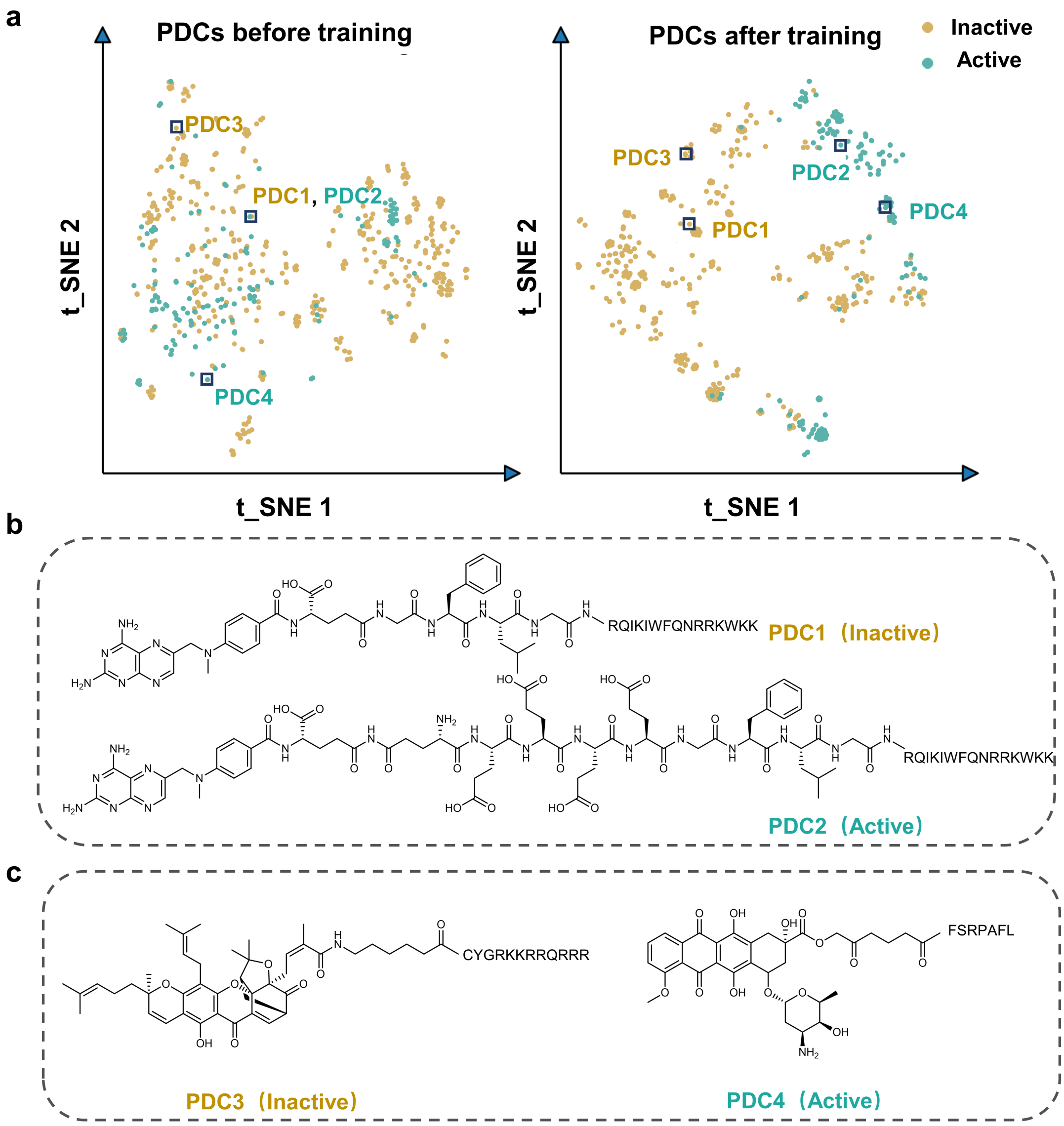

**Fig. 6 | Visualization analysis of PDCNet by t-SNE dimensionality reduction. a,** Spatial distribution of the dataset before and after training. **b**, Representation of the overlap of inactive PDC1 and active PDC2 in the pre-training space. **c**, Representation of the spatially unordered distribution of inactive PDC3 and active PDC4 in the pre-training space.

## Model interpretability analysis

We employed the SHAP (SHapley Additive exPlanations) algorithm to conduct interpretability analysis on the four input features of the PDCNet model. As shown in Fig. 7a, the contributions of the three structural features (peptide, linker, and payload) are generally comparable, with values of 0.79, 0.7, and 0.8, respectively. This demonstrates that the PDCNet model comprehensively integrated the combined effects of these three components on PDC activity during the modeling process, avoiding dependency bias towards any single module. These findings also indirectly verify the multiplicity of the SARs in PDCs, highlighting that the peptide, linker, and payload collectively determine the overall therapeutic performance.

We selected three peptides of appropriate lengths. The first two peptides share the same linker and payload, differing only in their peptide sequences, while the third peptide was chosen entirely at random. Fig. 7b depicts the relative importance of amino acids for each peptide sequence, as determined by attention scores. The results show that positively charged amino acids, such as lysine and arginine, have higher attention scores, indicating that the model recognizes the significance of basic residues for PDC activity. This observation aligns with the hypothesis that their cationic properties enhance cell targeting or membrane permeability[10, 36]. Notably, amino acids proximal to the linker attachment displayed highest attention scores, suggesting that the PDCNet have identified the positions where peptides are connected to linkers.

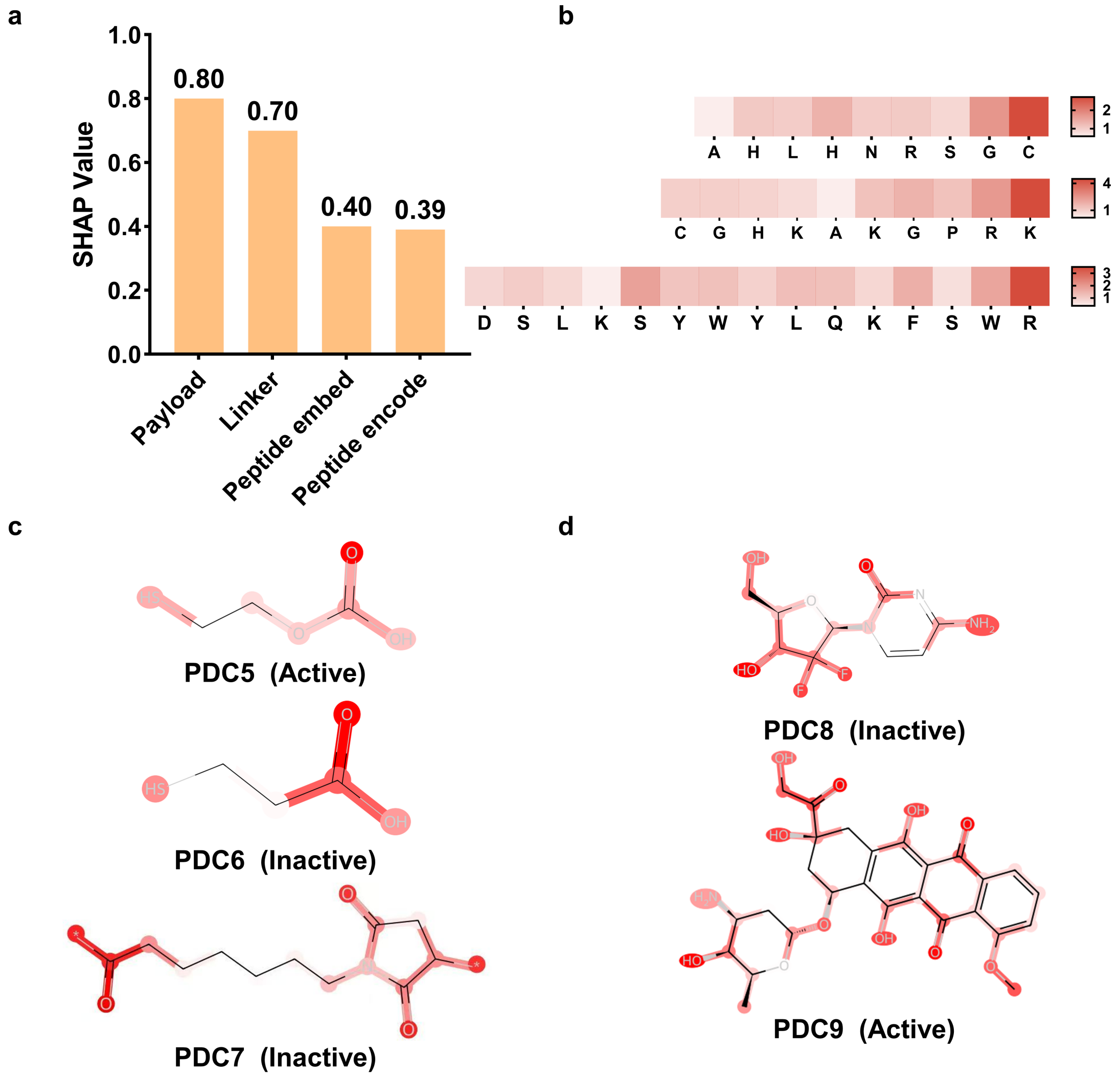

**Fig. 7 | Model interpretability analysis for PDCNet. a**, Feature contribution interpretation via SHAP values. **b**, Attention scores of peptides. **c**, Attention scores of linkers. **d**, Attention scores of payloads. Structural details for **PDCs5-9** are provided in Supplementary Fig. S2.

We selected three PDCs with identical peptides and payloads but different linkers to analyze the model's capacity to recognize linker-specific features. In Fig. 7c, color intensity denotes the magnitude of the attention weights, with darker colors indicating higher model attention. For the active PDC (PDC5, Fig. S2a), the model focuses more attention on the key functional group regions within the linker structure, particularly near the carboxyl (-COOH) and thiol (-SH) groups. This demonstrates that the model can identify structural features highly associated with activity and assign them greater importance, thereby significantly influencing the final activity prediction. Conversely, the linkers of the inactive PDCs (PDC6 and PDC7, Fig. S2b-c) exhibit different attention distribution characteristics: the attention is more dispersed, and key structural sites such as functional groups or regions with high electron density do not receive significantly higher weights. This suggests that the model fails to capture effective structural activity information from the linkers of inactive molecules or considers these structures to have a smaller contribution to the overall activity of PDC.

Additionally, we selected a pair of PDCs with opposing activities that differ solely in their payloads to analyze the model's capability to recognize payload-specific features. The payload of active PDC8 (Fig. S2e) contains multiple hydroxyl (-OH) and carbonyl (C=O) functional groups, which play a crucial role in intermolecular interactions and may enhance the binding affinity of the molecule to its target. Moreover, these hydroxyl groups can form hydrogen bonds with water molecules, thereby increasing its water solubility and facilitating the conjugation of PDC[37]. In contrast, the fluorine atoms in the payload of inactive PDC9 (Fig. S2f) may enhance the lipophilicity of the molecule, potentially leading to non-specific binding and accumulation within the biological system, thereby reducing its activity.

**Independent testing based on a new external dataset in real-word application scenarios**

The aforementioned results have confirmed the robustness and stability of PDCNet,

we further evaluate its generalization capability in real-word application scenarios. To this end, awe collected an external dataset consisting of 21 novel PDCs from recently published literature and patents to conduct independent testing[38, 39, 40, 41, 42, 43, 44, 45, 46, 47]. Details of these PDCs are provided in Supplementary Table 1.

To demonstrate the dissimilarity between the peptides and drugs conjugates (PDCs) in the external dataset and the benchmark PDC dataset, we employed distinct similarity calculation methods based on the characteristics of peptides and small molecules. For peptides, we utilized global sequence similarity alignment from Biopython, comparing each peptide in the new PDCs with those in the original dataset pairwise, normalizing the alignment scores, and finally outputting the maximum similarity score. For linkers and payloads, we adopted the Tanimoto similarity calculation method based on ECFP_4 fingerprints from RDKit, comparing each linker/payload in the new PDCs with those in the original dataset pairwise and outputting the maximum similarity score. Based on these similarity scores, we calculated the harmonic mean of the similarity scores for each new PDC. The results are shown in Table 2, with the minimum harmonic mean similarity score of the external dataset being 0.3600. This indicates the novelty of the external dataset, which can be used to validate the reliability and generalization ability of the PDCNet model.

## Table 2 | Performance of the PDCNet on an independent external dataset.

| No. | Peptide | Linker[1] | Payload[2-3] | Similarity scor[4] | Bioactivity | Label | Predictive score |
|---|---|---|---|---|---|---|---|
| 1 | YRSRKYSSWYVALKRLPET GGG | 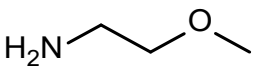 | Dinitrophenyl | 0.3600 | $IC_{50}$ = 12.5-25 μM | 0 | 0.0048 |
| 2 | YRSRKYSSWYVALKRLPET GGG | 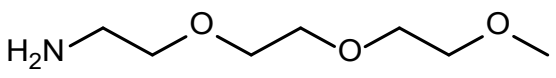 | Dinitrophenyl | 0.4112 | $IC_{50}$ = 25-50 μM | 0 | 0.0308 |
| 3 | RPPR | 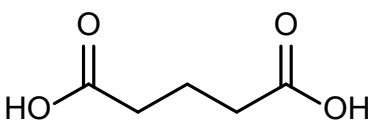 | Paclitaxel | 0.8182 | $IC_{50}$ = 0.26 μM | 1 | 0.9693 |
| 4 | YHWYGYTPERVI | 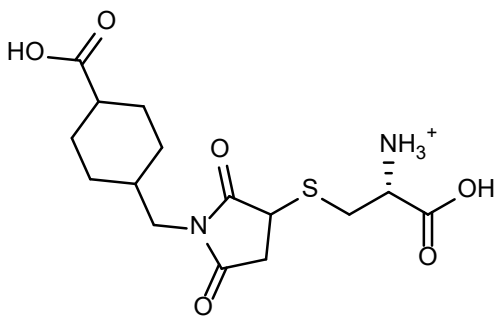 | Doxorubicin | 0.8003 | $IC_{50}$ = 2.3 μM | 0 | 0.0494 |
| 5 | KGDEVD | 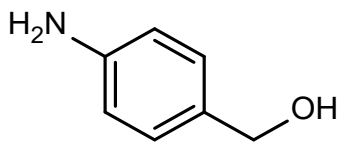 | Docetaxel | 0.7200 | $IC_{50}$ = 0.030 μM | 1 | 0.990 |
| 6 | GSS | 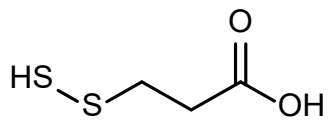 | Docetaxel | 0.8572 | $IC_{50}$ = 121.1-174.1 μM | 0 | 0.5446 |
| 7 | RGDC | 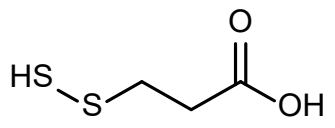 | Docetaxel | 0.9000 | $IC_{50}$ = 41.4-87.1 μM | 0 | 0.55 |
| 8 | FVDLKCIANCNSIFGK | 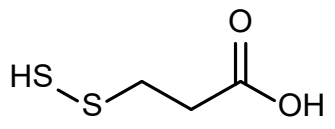 | Podophyllotoxin | 0.6206 | $IC_{50}$ = 0.22-0.88 μM | 1 | 0.9548 |
| 9 | CHVPGSYIC | 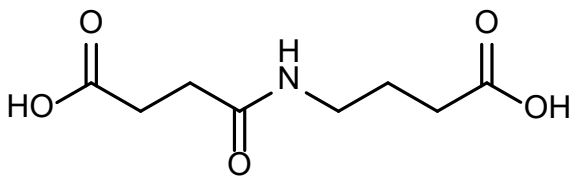 | SN-38 | 0.7068 | $IC_{50}$ = 0.9 μM | 1 | 0.9575 |
| 10 | KPSSPPEEK | 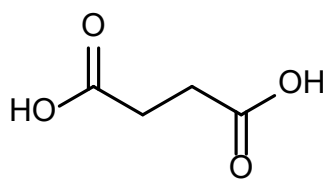 | Tubulin inhibitor 5B | 0.4941 | $IC_{50}$ = 0.23 μM | 1 | 0.0013 |
| 11 | GCTKSIPPICSPGAK | 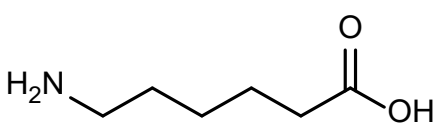 | Verteporfin | 0.4875 | In vivo study | 1 | 0.9891 |
| 12 | GCGGPLYKKIIKKLLESGG AGGAPLYKKIIKKLCES | 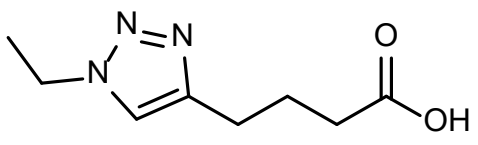 | Camptothecin | 0.5302 | $IC_{50}$ > 2 μM | 0 | 0.0166 |
| 13 | GCGGPLYKKIIKKLLESGG<br>AGGAPLYKKIIKKLCES | 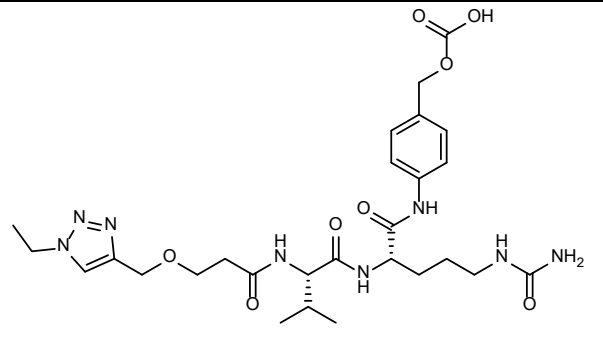 | Camptothecin | 0.5744 | $IC_{50}$ = 1.21 μM | 0 | 0.0343 |
| 14 | GCGGPLYKKIIKKLLESGG<br>AGGAPLYKKIIKKLCES | 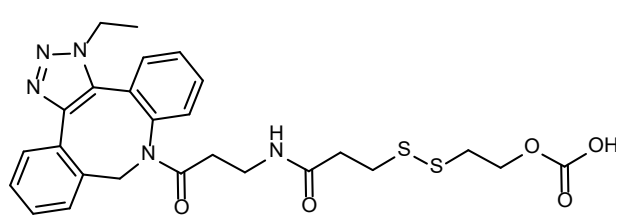 | Camptothecin | 0.4793 | $IC_{50}$ = 0.44 μM | 1 | 0.7150 |
| 15 | GGCGGAPLYKKIIKKLLES<br>GGCGGAPLYKKIIKKLLES | 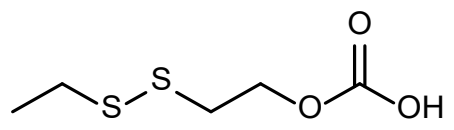 | Camptothecin | 0.5902 | $IC_{50}$ = 0.36 μM | 1 | 0.0759 |
| 16 | RGDFK | 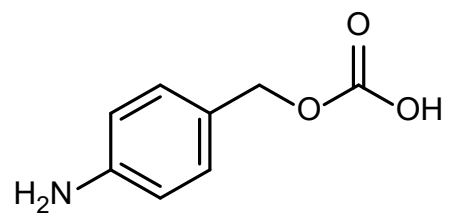 | Pomalidomide | 0.4925 | In vivo study | 1 | 0.8217 |
| 17 | FFRFKFRFK | 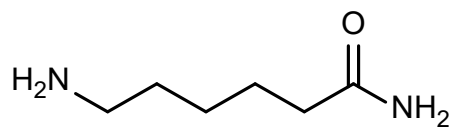 | Lonidamine | 0.5507 | $IC_{50}$ = 14.22 μM | 0 | 0.0001 |
| 18 | FFRFKFRFK | 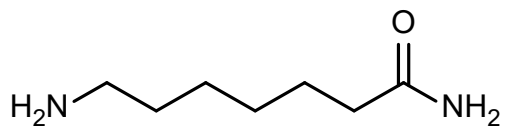 | Lonidamine | 0.5159 | $IC_{50}$ = 29.76 μM | 0 | 0.0013 |
| 19 | FFRFKFRFK | 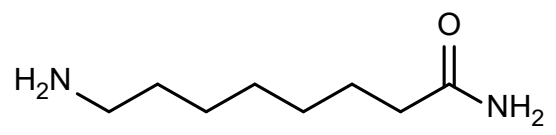 | Lonidamine | 0.5067 | $IC_{50}$ = 24.23 μM | 0 | 0.0020 |
| 20 | FFRFKFRFK | 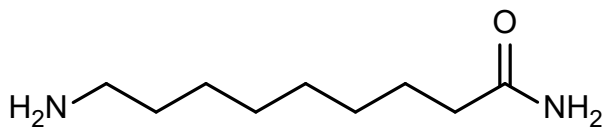 | Lonidamine | 0.5022 | $IC_{50}$ = 19.49 μM | 0 | 0.0035 |
| 21 | FFRFKFRFK | 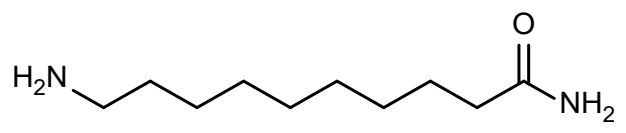 | Lonidamine | 0.4978 | $IC_{50}$ = 21.91 μM | 0 | 0.006 |

[1]SMILES of the linkers are in Table S1.

[2]SMILES of the payloads are in Table S1.

[3]The detailed structures of payloads are in Supplementary Fig. S3.

[4]The similarity score is defined as the harmonic mean of the similarity values. It determines the average level of similarity among a collection of data points by computing the reciprocal of the arithmetic average of the reciprocals of the similarity scores.

## Discussion

In this study, we proposed PDCNet, a unified deep learning framework based on the standardized PDC benchmark dataset for predicting the activity of PDCs. By systematically integrating peptide sequences, linkers, and small molecule payload structures, PDCNet demonstrated excellent performance in predicting the activity of various PDC samples. The construction of the PDC benchmark dataset played a crucial role, providing high-quality and standardized data resources to support model training, validation, and objective performance evaluation.

Although PDCNet was primarily trained on PDC activity data in the field of oncology and demonstrated good predictive ability in screening anti-cancer PDCs, its unified and general architecture design and feature extraction strategy mean that the model is not limited to oncological indications. If data on cell/target activity and corresponding PDC structures related to other diseases (such as infections, inflammation, and metabolic diseases) are introduced, PDCNet can also perform cross-disease activity prediction, showing good scalability and application potential. PDCNet has a unified and scalable framework that can effectively handle multimodal input data and can be flexibly expanded as the dataset continues to grow. This lays the foundation for its role as a basic tool for PDC activity research and helps to achieve PDC activity prediction tasks across targets and multiple disease fields in the future.

Looking to the future, we plan to further refine and expand this research work in the following directions. First, we will continue to collect and integrate the latest published PDC structures and their bioactivity data to expand the PDC benchmark dataset, thereby enhancing the model's generalization ability and scope of application. Second, to more comprehensively capture the biological context information that affects PDC activity, we will introduce more cellular biological features, such as target features, to achieve more accurate activity prediction. In addition, we also plan to combine generative deep learning techniques to develop an intelligent PDC molecule generation platform, realizing an integrated process from structural design to activity prediction, thus accelerating the innovation and optimization of novel PDC molecules. In summary, PDCNet, supported by a high-quality dataset and characterized by its unified and flexible model architecture, provides robust support for the computational prediction and rational design of PDCs. With the continuous expansion of data scale and feature dimensions, PDCNet is expected to play a more significant role in the future development of PDC drugs.

## Methods

### PDCs benchmark dataset construction

As shown in Fig. 1b, the benchmark dataset employed in the PDCNet was derived from two distinct sources: published academic papers and the PDCdb[48]. Firstly, a preliminary search was conducted in PubMed (https://pubmed.ncbi.nlm.nih.gov/) using the keyword 'Peptide drug conjugation', resulting in the manual collection of structure and activity data for anticancer PDCs, amounting to a total of 767 entries. Additionally, we collected 708 anticancer PDCs from the PDCdb. The two datasets were subsequently combined and processed as follows: 1) duplicate entries were removed; 2) PDCs with complete structure information were retained, including peptide sequences, SMILES of linkers and SMILES of payloads; 3) the data were standardized, including the removal of data containing non-standard amino acids; 4) unifying the units of biological activity data at the cellular level to μM, including $IC_{50}$, $EC_{50}$, and $GI_{50}$ (e.g., nM, pM, etc.); 5) if a PDC has multiple bioactivity data, retaining and selecting the minimum testing value as the final value. Through the above processing, a total of 834 unique structures of PDCs were finally obtained.

Subsequently, the data were categorized as follows: 1) if the PDCs were in the research status of marketing, clinical phase I, II and III, they were categorized as active; 2) if the PDCs were under investigation and animal experiments were conducted, they were also labelled as active; 3) if the PDCs were under investigation and only cellular experiments were performed, they were categorized as active based on $IC_{50}$, $EC_{50}$, and $GI_{50}$ values of <=1 μM, otherwise they were categorized as inactive; 4) if the PDCs were under investigation and no activity experiments have been performed, they are also labelled as negative. Details of the PDCs benchmark dataset are provided in Supplementary Table 2. The benchmark dataset was then randomly divided into three non-overlapping subsets, with 80% allocated for training, 10% for validation, and 10% for testing.

**PDCNet architecture**

The architecture of PDCNet is shown in Fig. 1c. In general, it contains three main modules: peptide features extraction module, linker and payload features extraction module, and a multi-channel features fusion and prediction module. Details of these three modules are described as the following.

**Peptide features extraction module**

To extract peptide features at PDCs, both amino acid- and peptide- level feature extractions were conducted (Fig. 1c, left). At the amino acid level, four encodings were performed for each amino acid following the input of the sequence containing one-hot encoding (20 dimensions), BLOSUM62 encoding (20 dimensions), positional encoding (20 dimensions), and Z-scale encoding (5 dimensions). This resulted in the amino acid encoding being transformed into a 65-dimensional feature vector $x_t \in R^{65}$. The entire peptide sequence can thus be represented as a time-series matrix $X_{Peptide}$, which is used for subsequent sequence modeling. Next, the time-series matrix is fed into the BiLSTM, generating a hidden state vector $h_t \in R^{d_h}$ at each time step, thereby forming the hidden state output sequence H. Subsequently, the self-attention mechanism is introduced, where the output sequence H from the BiLSTM is used simultaneously as the Query (Q), Key (K), and Value (V) for the attention computation. This ultimately results in a 512-dimensional peptide local feature vector $t_1 \in R^{512}$. The formula is as follows:

$$X_{Peptide} = [x_1, x_2, \cdots, x_T],\ x_t \in R^{65} \quad (1)$$

$$\mathrm{H} = \mathrm{BiLSTM}(X_{Peptide}) = [h_1, h_2, \cdots, h_T], h_t \in R^{d_h} \quad (2)$$

$$t_1 = Attention(H, H, H) = softmax\left(\frac{HH^T}{\sqrt{d_h}}\right)H \quad (3)$$

Here, $T$ represents the length of the peptide sequence, and $d_h$ denotes the dimension of the hidden layer in the BiLSTM.

For another, Evolutionary Sequence Model 2 (ESM-2), a protein language model developed by the Meta AI team, was employed to extract peptide-level features[49]. This model comprises six distinct models of varying complexity, with parameter counts spanning from 8 million parameters to 15 billion parameters. The models in question range from a 48-layer structure with 5120 output embeddings to smaller models comprising just six layers and 320 output embeddings. We selected the pLM with 640 output embeddings. Ultimately, amino acid-level and peptide-level features of peptide item of PDC are integrated with molecular features of linker and payload items of PDC underneath to generate predictions. Molecular characterizations for linker and payload of PDC are described as follows.

**Linker and payload features extraction module**

Recently, FG-BERT[50], a generic self-supervised molecular representation learning model based on functional groups, has been developed in our group and can be fine-tuned to downstream drug discovery related tasks, such as molecular properties prediction. Briefing, the model was trained on a corpus of approximately 1.45 million unlabeled molecules. Initially, the input molecular SMILES were converted to molecular graphs using the RDKit. Then, functional groups were masked in the molecular graphs to facilitate effective large-scale pre-training. Finally, chemical structure and semantic information were comprehensively mined from the corpus to enable the learning of useful molecular representations. In this study, we reasonably attribute the linkers and payloads of PDCs to small molecule terms based on their structures, and use a fine-tuned FG-BERT model to obtain feature extraction of the linkers and payloads (Fig. 1c, right). Given that the FG-BERT model exhibits superior performance in the evaluation of molecular properties, this enables accurate learning of useful feature representations of linkers and payloads of PDCs. As a consequence of this process, the small molecule structures of the linkers and payloads are transformed into 256-dimensional feature vectors, which are then subjected to subsequent processing.

**Multi-channel features fusion and prediction module**

According to the well-designed PDCNet architecture, the peptide portion is obtained through dual channels with feature vectors $t1$ and $t2$ at the amino acid- and peptide- levels, respectively. The linker and payload are then extracted by the FG-BERT model, resulting in feature vector $x1$ for the linker and feature vector $x2$ for the payload. In the multi-channel feature and prediction module, these feature vectors are then merged using the concat function, whose formula is delineated as:

$$x = concat(x_1, x_2, t_1, t_2) \quad (4)$$

The final task of feature integration is executed by a fully connected layer, as shown in the following equation:

$$\hat{y} = Wx + b \quad (5)$$

Where, $\hat{y}$ is the output vector, W is the weight matrix, $x$ is the input vector and $b$ is the bias vector.

**Model training**

PDCNet model has been developed on a Python and TensorFlow-based framework[51], which has been combined with a well-designed set of training strategies.

In the training phase, the dataset is divided into three parts: a training set, a validation set, and a test set. The training set is used to build the PDCNet model, the validation set is used to optimize the hyperparameters of PDCNet model, and the test set is employed to test the accuracy of the model. Three distinct random seeds (Fig. S4) are employed to facilitate the stochastic partitioning of the dataset, resulting in an 8:1:1 ratio. Upon loading the data, the model assumes a shared structure with independent output layers. PDCNet employs an early-stopping strategy to enhance the training process and circumvent overfitting, thereby augmenting its generalization capacity. The maximum number of training rounds was set to 50, with a tolerance of 10 for early stopping. In each training round, the model assesses and records AUC value of the validation set. In the event that the AUC value of the validation set does not demonstrate improvement after 10 consecutive rounds of training, the training process will be terminated prematurely and the evaluation will be conducted on the test set. At the conclusion of each training cycle, the AUC value of the current validation set is evaluated to ascertain whether it surpasses the previous optimal record. Subsequently, the current model parameters are saved as the new optimal model. In the event that the aforementioned condition is not met, the Early Stop Monitor counter is incremented by one. Upon reaching a total of 10, the training process is terminated and the test set is subsequently evaluated using the saved optimal model parameters. In the event that the early stop counter does not reach 10 following 50 rounds of training, the training is also terminated and the test set is evaluated instead. The early-stop strategy has the dual benefit of preventing model overfitting during training while also conserving computational resources and time. In addition, Hyperopt and Python-based Bayesian optimization techniques are employed as a means of optimizing and tuning the hyperparameters of the model. A total of 10 searches were conducted within a predefined parameter space with the objective of identifying the optimal combination of hyperparameters for achieving the best possible model performance.

**Baseline models construction and evaluation metrics**

With the aim of facilitating a comparative analysis of our PDCNet model with the baseline methods, we constructed eight baseline models. Notably, although these serve as baseline models for comparative analysis, like PDCNet model, they represent the first established models for predicting PDC anti-cancer activity. To our knowledge, no computational models have been previously developed specifically for PDC efficacy prediction in oncology research. In this study, four traditional ML algorithms, including RF[52], SVM[53], LR[54], and XGBoost[55]were unitilzed to constructed these baseline models. Detials of these methods and models are briefly described in Supplementary note. The first three models are integrated into the scikit-learn package, whereas the XGBoost model is incorporated into the XGBoost Python library. Two molecular fingerprinting techniques are employed to characterise the linkers and payloads of the PDCs, namely a 166-bit Molecular Access Communication System (MACCS) key[56] and a 1024-bit Morgan fingerprints[57]. The RF, SVM, LR, and XGBoost-based models synthesise features derived from molecular fingerprints and peptide chain sequences of linkers and payloads. Finally, PDCNet and these baseline models, were trainedd on a GPU (NVIDIA Corporation GV100GL, Tesla V100 PCIe 32 GB) and CPU (Intel(R) Xeon(R)

Silver 4216 CPU @2.10 GHz).

To comprehensively evaluate the predictive performance of PDCNet, ten standard metrics were employed as follows: accuracy (ACC), area under the receiver operating characteristic curve (AUC), F1 score, sensitivity (SE), specificity (SP), matthews correlation coefficient (MCC), precision-recall area under the curve (PRAUC), Positive predictive value (PPV) and negative predictive value (NPV). It is notable that the enhanced predictive power of the model corresponds to higher values of these metrics. The calculation of these evaluation metrics is outlined herewith:

$$\mathrm{ACC} = \frac{\mathrm{TP} + \mathrm{TN}}{TP + TN + FP + FN} \tag{6}$$

$$\mathrm{AUC} = \int_0^1 TPR(fpr)\, d(fpr) \tag{7}$$

$$\mathrm{F1} = \frac{2 \times TP}{2 \times TP + FN + FP} \tag{8}$$

$$SE = \frac{TP}{TP + FN} \tag{9}$$

$$SP = \frac{TN}{TP + FP} \tag{10}$$

$$MCC = \frac{TP \times TN - FN \times FP}{\sqrt{(TP + FN) \times (TP + FP) \times (TN + FN) \times (TN + FP)}} \tag{11}$$

$$BA = \frac{TPR + TNR}{2} = \frac{SE + SP}{2} \tag{12}$$

$$\mathrm{PRAUC} = \int_0^1 PR(R^{-1}(t))dt \tag{13}$$

$$PPV = \frac{TP}{TP + FP} \tag{14}$$

$$NPV = \frac{TN}{TN + FN} \tag{15}$$

where, TP is true positive, FP is false positive, FN is false negative, TN is true negative, TPR is true positive rate and TNR is true negative rate.

**Data availability**

The source code of PDCNet and associated data preparation scripts are available at GitHub (https://github.com/idrugLab/PDCNet). The PDCs data used in the present study are freely available in PDCdb (https://idrblab.org/PDCdb/). The remaining data or questions regarding this study are available to the corresponding author upon request (Ling Wang: lingwang@scut.edu.cn).

**Code availability**

The source code of PDCNet and associated data preparation scripts are available at github (https://github.com/idruglab/PDCNet). The optimized PDCNet model is also provided.

**Acknowledgements**

This work was supported by the Natural Science Foundation of Guangdong Province (No. 2023B1515020042), Natural Science Foundation of China (22267003), the Guizhou Provincial Excellent Young Talents Plan (YQK[2023]030), and the Excellent Young Talents Plan of Guizhou Medical University ([2022]102). We are also grateful for the support from HPC Platform of South China University of Technology.

**Author contributions**

Y.L. collected, annotated, and processed the PDCs data, helped train the PDCNet model, and wrote the manuscript; J.H. developed and trained the network model; Y.Z., J.Z.,

C.L., X.L., C.W., and Y.P. helped for data collection, annotation and processing; L.W. designed the entire project, revised the manuscript and solved the problems encountered.

**Competing interests**

The authors declare no competing interests.

**Additional information**

**Supplementary information**

The online version contains supplementary material available at

**Correspondence** and requests for materials should be addressed to Ling Wang.

**References**

1. Strebhardt K&Ullrich AJNRC. Paul Ehrlich's magic bullet concept: 100 years of progress. *Nat Rev Cancer* **8**, 473-480 (2008).
2. Valent P*, et al.* Paul Ehrlich (1854-1915) and His Contributions to the Foundation and Birth of Translational Medicine. *J Innate Immun* **8**, 111-120 (2016).
3. Beck A, Goetsch L, Dumontet C&Corvaïa NJNrDd. Strategies and challenges for the next generation of antibody–drug conjugates. *Nat Rev Drug Discovery* **16**, 315-337 (2017).
4. Su Z*, et al.* Antibody-drug conjugates: Recent advances in linker chemistry. *Acta pharmaceutica Sinica B* **11**, 3889-3907 (2021).
5. Smaglo BG, Aldeghaither D&Weiner LM. The development of immunoconjugates

for targeted cancer therapy. *Nat Rev Clin Oncol* **11**, 637-648 (2014).

6. Chari RV, Miller ML&Widdison WCJACIE. Antibody–drug conjugates: an emerging concept in cancer therapy. *Angew Chem Int Ed Engl* **53**, 3796-3827 (2014).

7. Izzo D*, et al.* Innovative payloads for ADCs in cancer treatment: moving beyond the selective delivery of chemotherapy. *Ther Adv Med Oncol* **17**, 17588359241309461 (2025).

8. Colombo R, Tarantino P, Rich JR, LoRusso PM&de Vries EGE. The Journey of Antibody-Drug Conjugates: Lessons Learned from 40 Years of Development. *Cancer Discov* **14**, 2089-2108 (2024).

9. Zhang B*, et al.* Recent Advances in Targeted Cancer Therapy: Are PDCs the Next Generation of ADCs? *J Med Chem* **67**, 11469-11487 (2024).

10. Rizvi SFA, Zhang L, Zhang H&Fang Q. Peptide-Drug Conjugates: Design, Chemistry, and Drug Delivery System as a Novel Cancer Theranostic. *ACS Pharmacol Transl Sci* **7**, 309-334 (2024).

11. Rizvi SFA, Zhang H&Fang Q. Engineering peptide drug therapeutics through chemical conjugation and implication in clinics. *Med Res Rev* **44**, 2420-2471 (2024).

12. Laakkonen P&Vuorinen K. Homing peptides as targeted delivery vehicles. *Integr Biol* **2**, 326-337 (2010).

13. Hoppenz P, Els-Heindl S&Beck-Sickinger AG. Peptide-Drug Conjugates and Their Targets in Advanced Cancer Therapies. *Front Chem* **8**, 571 (2020).

14. Halmos G, Nagy A, Lamharzi N&Schally AV. Cytotoxic analogs of luteinizing hormone-releasing hormone bind with high affinity to human breast cancers. *Cancer Lett* **136**, 129-136 (1999).

15. Wang L, Chen H, Wang F&Zhang X. The development of peptide-drug conjugates (PDCs) strategies for paclitaxel. *Expert Opin Drug Deliv* **19**, 147-161 (2022).

16. Zhang Y*, et al.* Development of peptide paratope mimics derived from the anti-ROR1 antibody and long-acting peptide-drug conjugates for targeted cancer therapy. *J Med Chem* **67**, 10967-10985 (2024).

17. Muratspahić E*, et al.* Design and structural validation of peptide-drug conjugate ligands of the kappa-opioid receptor. *Nat Commun* **14**, 8064 (2023).

18. Zhu Y, Zhang Y, Li X&Wang L. 3MTox: A motif-level graph-based multi-view chemical language model for toxicity identification with deep interpretation. *J Hazard Mater* **476**, 135114 (2024).

19. Ai D*, et al.* DEEPCYPs: A deep learning platform for enhanced cytochrome P450 activity prediction. *Front Pharmacol* **14**, 1099093 (2023).

20. Cai H, Zhang H, Zhao D, Wu J&Wang L. FP-GNN: a versatile deep learning architecture for enhanced molecular property prediction. *Brief Bioinform* **23**, (2022).

21. Zhao D*, et al.* Highly Accurate and Explainable Predictions of Small-Molecule Antioxidants for Eight In Vitro Assays Simultaneously through an Alternating Multitask Learning Strategy. *J Chem Inf Model* **64**, 9098-9110 (2024).

22. Zhu W, Zhang Y, Zhao D, Xu J&Wang L. HiGNN: A Hierarchical Informative

Graph Neural Network for Molecular Property Prediction Equipped with Feature-Wise Attention. *J Chem Inf Model* **63**, 43-55 (2023).

23. Wu J*, et al.* Large-scale comparison of machine learning methods for profiling prediction of kinase inhibitors. *J Cheminform* **16**, 13 (2024).
24. Lin M*, et al.* MalariaFlow: A comprehensive deep learning platform for multistage phenotypic antimalarial drug discovery. *Eur J Med Chem* **277**, 116776 (2024).
25. Zhang Y*, et al.* Self-awareness of retrosynthesis via chemically inspired contrastive learning for reinforced molecule generation. *Brief Bioinform* **26**, (2025).
26. Tamura K, Stecher G&Kumar S. MEGA11: Molecular Evolutionary Genetics Analysis Version 11. *Mol Biol Evol* **38**, 3022-3027 (2021).
27. Saitou N&Nei M. The neighbor-joining method: a new method for reconstructing phylogenetic trees. *Mol Biol Evol* **4**, 406-425 (1987).
28. Heh E*, et al.* Peptide Drug Conjugates and Their Role in Cancer Therapy. *Int J Mol Sci* **24**, (2023).
29. Wang M*, et al.* Peptide-drug conjugates: A new paradigm for targeted cancer therapy. *Eur J Med Chem* **265**, 116119 (2024).
30. Chavda VP, Solanki HK, Davidson M, Apostolopoulos V&Bojarska J. Peptide-Drug Conjugates: A New Hope for Cancer Management. *Molecules* **27**, (2022).
31. Landrum G. RDKit: Open-source cheminformatics from machine learning to chemical registration. *Abstr Pap Am Chem Soc* **258**, 1 (2019).
32. Thölke P*, et al.* Class imbalance should not throw you off balance: Choosing the right classifiers and performance metrics for brain decoding with imbalanced data.

*Neuroimage* **277**, 120253 (2023).

33. Boughorbel S, Jarray F&El-Anbari M. Optimal classifier for imbalanced data using Matthews Correlation Coefficient metric. *PLoS ONE* **12**, 17 (2017).
34. Pang Y, Wang Z, Jhong JH&Lee TY. Identifying anti-coronavirus peptides by incorporating different negative datasets and imbalanced learning strategies. *Brief Bioinform* **22**, 1085-1095 (2021).
35. Van Der Maaten L&Hinton G. Visualizing Data using t-SNE. *J Mach Learn Res* **9**, 2579-2605 (2008).
36. Guidotti G, Brambilla L&Rossi D. Cell-Penetrating Peptides: From Basic Research to Clinics. *Trends Pharmacol Sci* **38**, 406-424 (2017).
37. McCombs JR&Owen SC. Antibody drug conjugates: design and selection of linker, payload and conjugation chemistry. *AAPS J* **17**, 339-351 (2015).
38. Li X*, et al.* Chemoenzymatic Synthesis of DNP-Functionalized FGFR1-Binding Peptides as Novel Peptidomimetic Immunotherapeutics for Treating Lung Cancer. *J Med Chem* **67**, 15373-15386 (2024).
39. Than PP, Yao SJ, Althagafi E&Kaur K. A Conjugate of an EGFR-Binding Peptide and Doxorubicin Shows Selective Toxicity to Triple-Negative Breast Cancer Cells. *ACS Med Chem Lett* **16**, 109-115 (2025).
40. Sun G*, et al.* Design and synthesis of isatin derivative payloaded peptide-drug conjugate as tubulin inhibitor against colorectal cancer. *Eur J Med Chem* **285**, 117276 (2025).
41. Han Z*, et al.* Design, synthesis and activity evaluation of reduction-responsive

anticancer peptide temporin-1CEa drug conjugates. *Bioorg Chem* **154**, 108103 (2025).

42. Cho YS*, et al.* Macropinocytosis-targeted peptide-docetaxel conjugate for bystander pancreatic cancer treatment. *J Control Release* **376**, 829-841 (2024).
43. Bazylevich A*, et al.* Novel Cyclic Peptide–Drug Conjugate P6-SN38 Toward Targeted Treatment of EGFR Overexpressed Non-Small Cell Lung Cancer. *Pharmaceutics* **16**, 1613 (2024).
44. Lepland A*, et al.* Peptide-Drug Conjugate for Therapeutic Reprogramming of Tumor-Associated Macrophages in Breast Cancer. *Adv Sci*, 2410288 (2025).
45. Li Y*, et al.* PTX-RPPR, a conjugate of paclitaxel and NRP-1 peptide inhibitor to prevent tumor growth and metastasis. *Biomed Pharmacother* **178**, 117264 (2024).
46. Li Q*, et al.* Reduction-Responsive RGD-Docetaxel Conjugate: Synthesis, In Vitro Drug Release and In Vitro Antitumor Activity. *Drug Dev Res* **86**, e70043 (2025).
47. Palombi IR*, et al.* Synthesis and Investigation of Peptide–Drug Conjugates Comprising Camptothecin and a Human Protein-Derived Cell-Penetrating Peptide. *Chem Biol Drug Des* **105**, e70051 (2025).
48. Sun X*, et al.* PDCdb: the biological activity and pharmaceutical information of peptide-drug conjugate (PDC). *Nucleic Acids Res* **53**, D1476-1485 (2024).
49. Lin ZM*, et al.* Evolutionary-scale prediction of atomic-level protein structure with a language model. *Science* **379**, 1123-1130 (2023).
50. Li BS, Lin MJ, Chen TG&Wang L. FG-BERT: a generalized and self-supervised functional group-based molecular representation learning framework for properties prediction. *Brief Bioinform* **24**, 13 (2023).

51. Abadi M*, et al.* TensorFlow: A system for large-scale machine learning. *In 12th USENIX symposium on operating systems design and implementation (OSDI 16)*, 265-283 (2016).
52. Breiman L. Random forests. *Mach Learn* **45**, 5-32 (2001).
53. Cortes C&Vapnik V. Support-vector networks. *Mach Learn* **20**, 273-297 (1995).
54. Davis LJ&Offord KP. Logistic regression. *J Pers Assess* **68**, 497-507 (1997).
55. Chen T&Guestrin C. XGBoost: a scalable tree boosting system. In: *22nd ACM SIGKDD International Conference on Knowledge Discovery and Data Mining*). ACM (2016).
56. Durant JL, Leland BA, Henry DR, Nourse JGJJoci&sciences c. Reoptimization of MDL keys for use in drug discovery. **42**, 1273-1280 (2002).
57. Rogers D, Hahn MJJoci&modeling. Extended-connectivity fingerprints. **50**, 742-754 (2010).